\documentclass{article}
\usepackage[final]{neurips_2021}

\usepackage{amsmath}

\setcitestyle{square,numbers,comma}
\usepackage[utf8]{inputenc} 
\usepackage[T1]{fontenc}    
\usepackage{hyperref}       
\usepackage{url}            
\usepackage{booktabs}       
\usepackage{amsfonts}       
\usepackage{graphicx}
\usepackage{nicefrac}       
\usepackage{microtype}      
\usepackage{xcolor}         
\usepackage{booktabs}       
\usepackage{subcaption}
\usepackage{cleveref}

\title{Unsupervised Wildfire Change Detection based on Contrastive Learning}

\author{%
  Beichen Zhang$^{1}$, Huiqi Wang$^{2}$, Amani Alabri$^{3}$, Karol Bot$^{4}$, \\
  \textbf{Cole McCall$^{5}$,
  Dale Hamilton$^{5}$\thanks{Correspondence to: \href{mailto:dhamilton@nnu.edu}{dhamilton@nnu.edu}},
  Vít Růžička$^{6}$} \\
  ${}^1$University of Nebraska-Lincoln, \hfill ${}^2$University of California, Berkeley,\hfill ${}^3$Boston University, \\ ${}^4$Lisbon University,  ${}^5$Northwest Nazarene University, ${}^6$University of Oxford \\
}

\begin{document}

\maketitle

\begin{abstract}
    The accurate characterization of the severity of the wildfire event strongly contributes to the characterization of the fuel conditions in fire-prone areas, and provides valuable information for disaster response. The aim of this study is to develop an autonomous system built on top of high-resolution multispectral satellite imagery, with an advanced deep learning method for detecting burned area change. This work proposes an initial exploration of using an unsupervised model for feature extraction in wildfire scenarios. It is based on the contrastive learning technique \textit{SimCLR}, which is trained to minimize the cosine distance between augmentations of images. The distance between encoded images can also be used for change detection. We propose changes to this method that allows it to be used for unsupervised burned area detection and following downstream tasks. We show that our proposed method outperforms the tested baseline approaches.
\end{abstract}

\section{Introduction}

Wildfires are uncontrolled fires occurring in forests or grassland in wildland areas, which threaten and kill people, destroy urban and rural property, degrade air quality, ravage forest ecosystems, and contribute to global warming \cite{bot2022systematic}. Although fire has always been an integral part of many ecosystems, recent wildfires have shown an unprecedented spatial and temporal extent and go beyond controls of national firefighting resources \cite{moustakas2021minimal}. The connection between climate change and the increased risk of wildfires suggests a paradigm change in the co-existence of humans with natural catastrophes affecting the environment. The forest fires paradox has been highlighted in many studies. On the one hand, they may play an important ecological role by removing deadwood and opening space for new vegetation growth, and releasing nutrients into the soil, offering ecological niches for the proliferation of wildlife. On the other hand, when occurring at high intensity, forest fires lead to adverse environmental impacts such decreased soil quality and harm to life \cite{bot2022systematic}. The wildfires may further contribute to a decline in biodiversity and air quality, thus threatening forested landscapes.

\subsection{Application Context}

It is of fundamental importance to further develop the tools employed to assess wildfire events. In fire-prone areas, there is a high probability of occurrence of a next fire event, and the temporal development of the fuel conditions in that area will depend greatly on the severity of the previous wildfire. In this sense, a post-fire scenario becomes a pre-fire scenario for the next event. Considering the amount of climate-related disasters happening in the last decades and the probability of subsequent events, a new challenge for the research community arises – the need to develop more effective automated change detection methods that could work in various locations with acceptable results to detect the extension of a disaster such as a wildfire \cite{khelifi2020deep}.
In this context, our work aims to employ unsupervised machine learning methods associated with multispectral satellite images to assess the change detection of the burned areas.

\subsection{Background}


Overall, monitoring wildfires and their extent is essential for assessing ecological outcomes of fires and their spatial patterning as well as guiding efforts to mitigate or restore areas where ecological consequences are negative \cite{gallagher2021estimation}. Accurate burned area information is needed to assess the impacts of wildfires on people, communities, and natural ecosystems, as well as considerations of the fuel state for landscape management in fire scenarios \cite{gibson2020remote}. The fire severity assessment is crucial for predicting ecosystem response and prioritizing post-fire forest management strategies \cite{montorio2020unitemporal}. Besides, mapping burned areas also supports evacuation and accessibility to emergency facilities. This generalization is even more valuable for any use-case scenarios, including the organization of fire-fighting activities or civil protection \cite{florath2022supervised}.


Change detection is defined as identifying differences in a site's state or phenomenon by observing it at different times \cite{khelifi2020deep}. Over the years, many methods have been developed for analyzing remote sensing data, and newer techniques are still being developed. Timely and accurate change detection of Earth’s surface features provides the basis for evaluating the relationships and interactions between human and natural phenomena for better resource management. In general, change detection uses multi-temporal datasets to quantitatively analyze the temporal effects of the phenomenon \cite{asokan2019change}. A similar research question is explored in a previous study that used unsupervised change detection for a general disaster event detection \cite{ruzicka2021unsupervised} - we differ in the focus on wildfires as a disaster event type.

\section{Data}
\label{sec:data}

We used two sources for data in this work - the Sentinel-2 and PlanetScope \cite{planet} satellite imagery products. PlanetScope imagery has a spatial resolution of 3.7 meters in four multispectral bands (red, green, blue, and near infrared) with daily revisit. On the other hand, Sentinel-2 data is freely available with imagery of spatial resolution of 10 meters, a revisit time of 5 days, and is captured in 13 multispectral bands. One of the reasons why we were interested in using Sentinel-2 and PlanetScope imagery is because multispectral imagery has previously been used as inputs for determining fuel loads and burn severity \cite{hamilton2018evaluation}. Data acquisition and preprocessing steps are further described in Sections \ref{sec:data_acq} and \ref{sec:data_prep}.

\section{Methodology}
\label{sec:methodology}

Two types of models were developed, baseline and machine learning models. Figure \ref{fig:method} in the appendix shows the diagram of the proposed pipeline.
For the baseline models, we use the commonly used indices such as Normalized Difference Vegetation Index (NDVI) for PlanetScope and Normalized Burn Ratio (NBR) for Sentinel-2 data. The NDVI is calculated based on the red and the NIR bands. The NBR uses the NIR and the shortwave infrared (SWIR) bands \cite{escuin2008fire}.
For downstream unsupervised clustering experiments, we conduct Principal Component Analysis (PCA) for both products \cite{wold1987principal}. 


Our proposed method uses contrastive learning to obtain informative description of the data. Self-supervised learning is targeted for the task where there is not enough labeled data for use or annotating labels is too expensive. Contrastive learning aims to extract better representation for images by grouping similar samples closer and differing samples far from each other. More concretely, we use the SimCLR \cite{chen2020simple} method which maximizes agreement between two augmented versions of the same image, and proposes a simple framework to learn representations from unlabeled images based on heavy data augmentation. The original SimCLR uses the ResNet \cite{he2016deep} model in its encoding part, which is limited to only 3 bands (typically pretrained with RGB images) and is quite large. 
Therefore, we propose our framework \textbf{FireCLR}, where a small and simple convolutional neural network is used instead of the ResNet to extract features so that it works for the satellite imagery with multispectral bands and can be trained from scratch based on our data. Furthermore this gives us a flexible network architecture which could be trained with any number of bands, and variable resolutions of the used tiles.

\begin{figure}
    \centering
    \includegraphics[trim=100 120 100 100,clip,width=0.7\textwidth]{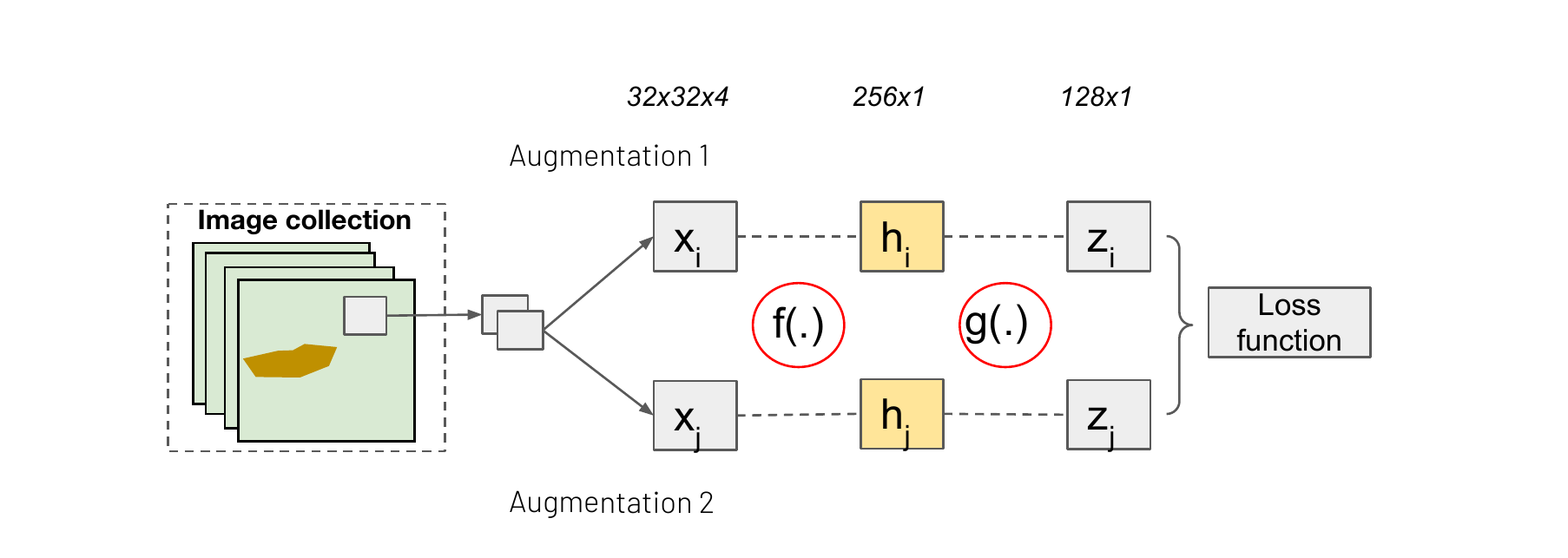}
    \caption{SimCLR structure, using our suggested augmentation scheme we denote it as the FireCLR model.\label{fig:fireclr}}
\vspace{-4mm}
\end{figure}

Same as SimCLR, our FireCLR includes the data augmentation, our own convolutional neural network backbone, a project head and the loss function (Figure \ref{fig:fireclr}). Data augmentation for image tiles includes: \textit{random crop}, \textit{gaussian blur}, \textit{random flip} and \textit{fixed rotation}. Importantly, we excluded any augmentation process which would change the color of the images since the changes caused by fires are sensitive to the reflectance. A customized convolution neural network is used here to receive an augmented image of shape (32,32,4) and outputs a 256-dimensional embedding vector $h$. Then a project head $g(.)$ is used to produce a final representation $z = g(h)$ from vector $h$. The Noise Contrastive Estimator (NCE) loss \cite{chen2020simple} is used as loss function.

We used the following formula to calculate the change map from the FireCLR model: \begin{equation}
\label{eqn:score_latentspace}
S(x) = d ( f( x_{t_1} ) , f( x_{t_2} ) )
\end{equation} Where $x_{t_1}$ and $x_{t_2}$ corresponds to the extracted tiles at the same location $x$ at times $t_1$ and $t_2$, $f(.)$ represents the learned feature encoder also highlighted on Figure \ref{fig:fireclr} and for $d(.)$ we used cosine distance. This score is calculated for each tile extracted from the sequence of satellite images. These scores can be aggregated and visualised as images shown in Figures \ref{fig:s2_results},  \ref{fig:ps_results}.

The FireCLR model was trained on 4 bands (RGB+NIR) of the Sentinel-2, on cumulative area of 320 km$^2$ captured at 8 temporal snapshots of the Mesa fire, finally tiled into 941,190 tiles of 32x32px. We have trained the model for 1024 epochs with an early stop on plateau.

When working with the PlanetScope data we trained the FireCLR model on 4 bands (RGB+NIR), on cumulative area of 2,937 km$^2$ captured at 43 temporal snapshots of the East Troublesome and Mcfarland fires, tiled into 4,382,607 tiles of 32x32px. We have trained the model for 1024 epochs with an early stop on plateau.

\section{Experiments}

In this work, the following model tests were done. First step was to obtain a representation, which would allow us to make comparisons to detect burned areas. For this task, we implemented baseline approaches to the task of unsupervised change detection, using two indices: differenced Normalized Burn Ratio (dNBR) and differenced Normalized Difference Vegetation Index (dNDVI). We also implemented a feature extractor pre-trained using the FireCLR framework.

Secondly, we applied k-means clustering to classify changes from the dNBR, dNDVI, and differenced FireCLR representations, obtained in the first step. In addition, we manually associated the clusters found based on the distance map with several classes connected with post fire activities. We note that the annotation effort of this last step is minimal.
We used the F1-score and the area under the precision recall curve (AUPRC) score as our major metrics to assess the model performance. 


We experimented with two different modes of training and evaluating the data. For one set of experiments, we used the same geographical location for both datasets but with different time snapshots. These time snapshots also include situations before and after the fire, however there is no need for manual annotation for this data. We consider these experiments valuable for evaluating how well machine learning models trained on very specific areas of interest perform. With enough unlabeled training data, we could keep many pre-trained and local models, specific to their areas of interest.

Secondly, we have run experiments with the aim towards global deployment. These models were trained on a set of multiple locations and with access to both pre-fire and post-fire data and evaluated on geographically and ecologically different locations. These results aim to obtain information about how well the learned machine learning models can generalize on new locations, given access to information about how fire damage looks in other locations. This mode of operation still doesn't require any manual labels.

\section{Results}
\label{sec:results}

From our first experiment we present the results when using the Sentinel-2 dataset in what we call local mode, where the same geographical location is used for training and evaluation (naturally using different time frames for both). This is a realistic scenario, as it is possible to obtain representative imagery from an area with a known fire. We do not require adding any labels to this data.

\begin{figure}[h]
    \centering
    \begin{subfigure}{0.30\textwidth}
    \includegraphics[width=\textwidth]{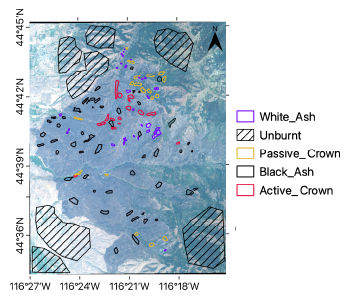}
    \caption{\label{fig:s2_gt}}
    \end{subfigure}\hspace{4mm}
    \begin{subfigure}{0.32\textwidth}
    \includegraphics[width=\textwidth]{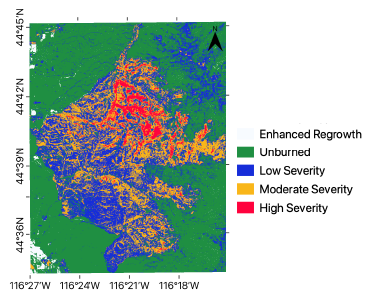}
    \caption{\label{fig:s2_base}}
    \end{subfigure}\hspace{4mm}
    \begin{subfigure}{0.30\textwidth}
    \includegraphics[width=\textwidth]{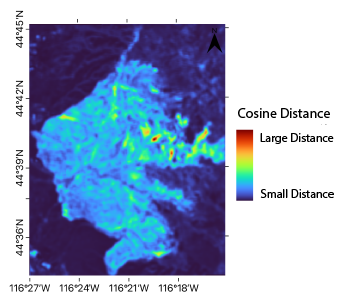}
    \caption{\label{fig:s2_our}}
    \end{subfigure}
    \caption{Results on the Sentinel-2 data used in the local mode at the Mesa fire. Ground truth in (a), the baseline method (dNBR) in (b) and our proposed FireCLR method in (c). Best seen on screen.\label{fig:s2_results}}
\vspace{-2mm}
\end{figure}

The \textbf{dNDVI baseline} based on the Sentinel-2 data achieves the AUPRC score of \textit{0.76}, while the \textbf{dNBR baseline} achieves the AUPRC score of \textit{0.95}, and our proposed \textbf{FireCLR} reaches the AUPRC score of \textit{0.96}. The results are visually shown on Figure \ref{fig:s2_results}. Compared to the ground truth Figure \ref{fig:s2_gt}, both the predictions based on the dNBR (Figure \ref{fig:s2_base}) and the cosine distance of the 128-dimensional latent space representation reveal the fire boundary well (Figure \ref{fig:s2_our}). The right colored pixels in Figure \ref{fig:s2_base} and \ref{fig:s2_our} indicate more significant changes, which could project the burned severity. The change map of the dNBR indicates larger areas in the upper-middle scene that were severely burned in the Mesa fire, while our proposed FireCLR indicates several noticeable spots from the upper-middle to the middle-right scene changed by the wildfire. 

\begin{figure}[h]
    \centering
    \begin{subfigure}{0.279\textwidth}
    \includegraphics[width=\textwidth]{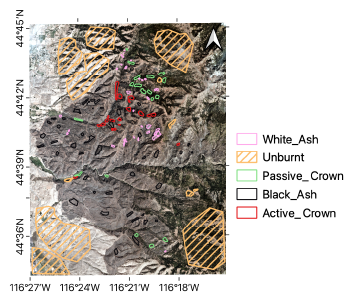}
    \caption{\label{fig:ps_gt}}
    \end{subfigure}\hspace{4mm}
    \begin{subfigure}{0.234\textwidth}
    \includegraphics[width=\textwidth]{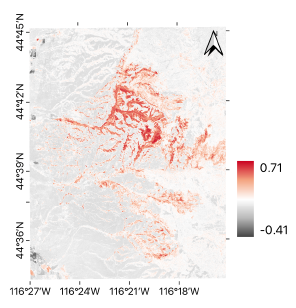}
    \caption{\label{fig:ps_base}}
    \end{subfigure}\hspace{4mm}
    \begin{subfigure}{0.252\textwidth}
    \includegraphics[width=\textwidth]{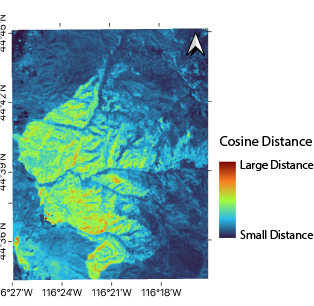}
    \caption{\label{fig:ps_our}}
    \end{subfigure}
    \caption{Results on the PlanetScope data used in the global mode at the Mesa fire. Ground truth in (a), the baseline method (dNDVI) in (b) and our proposed FireCLR method in (c). 
    \label{fig:ps_results}}
\vspace{-2mm}
\end{figure}

Secondly, we use the PlanetScope data in the global mode, where we select different geographical locations between the training and the evaluation sets (alongside different time frames). This scenario tests the capabilities of pre-training a model on a few locations with historical fire datasets and then using it on required areas with new fires.
The \textbf{dNDVI baseline} achieves the AUPRC score of \textit{0.67}, while our proposed \textbf{FireCLR} reaches the AUPRC score of \textit{0.80}. The results are visually shown on Figure \ref{fig:ps_results}. The fire perimeters predicted from the dNDVI and the proposed FireCLR model based on the PlanetScope data were shown in Figure \ref{fig:ps_base} and \ref{fig:ps_our}, which had lower contrast as compared to the Sentinel-2 data. The dNDVI map indicates broader significantly changed areas in the scene and primarily concentrated in the north. The cosine distance map of the FireCLR on the Planetscope data indicates similar patterns of the significantly changed areas.

We note that for these results the dNBR approach forms a strong baseline, but because it requires access to the SWIR bands that are available only in the Sentinel-2 data, we cannot always use it. In other scenarios we are limited to using simpler baselines, such as dNDVI. In the results on the Sentinel-2 dataset, our approach is maintaining the same performance as the strong baseline approach. When working with the PlanetScope data, we however see a boost of perfomance when using our method in comparison with the baseline approach. We also note that direct comparisons between the Sentinel-2 and the PlanetScope dataset are not possible in this scenario, as each was trained with different mode of operation (local or global).

In the appendix in section \ref{sec:additional} we also show additional results by running an unsupervised clustering algorithm on the obtained intermediate representations (either using PCA, or values from the trained FireCLR). The representations are hence summarized and compared to the burned severity: white ash and black ash. With the proposed \textbf{FireCLR}, the model trained on the PlanteScope data outperforms the dNBR approach based on the Sentinel-2 on classified black ash, while it performs worse on the white ash class. This is likely due to the different available data used in the two experiments (namely having access to the SWIR band). This points us to some interesting directions about including more bands in future experiments.

\section{Conclusion}

In conclusion, the proposed FireCLR model outperforms the baseline methods in both Sentinel-2 and PlanetScope datasets based on the AUPRC score, and shows mixed, but comparable results on downstream validation tasks.
We note that this work should serve as an initial exploration of this approach. In future research, we would like to run more experiments to closely compare the performance of using sensor data of different resolutions.

As an additional exciting future research direction, we would like to explore further adaptions of the SimCLR model. By constructing the positive pairs from longer temporal series of data, we can train the model to be invariant to the irrelevant natural changes. These improvements would enhance the post fire maps with knowledge of burn severity including both burn extent as well as biomass consumption. We note that this requires only minimal amount of supervision - selecting date ranges without any fires in the selected areas.
Detecting the fires burned severity achieved here with the trade-offs in spatial and spectral resolution of the two used data sets, namely Sentinel-2 and PlanetScope. We aim to comparing different satellite products and explore relevant to wildfire mapping. 
Finally, we would like to compare the results produced using contrastive learning against a variational auto-encoder approach of \cite{ruzicka2021unsupervised} to see if one algorithm is superior for mapping post fire effects.

We would like to also note the benefits of using unsupervised learning methods due to the ease they provide for future machine learning model adaptation on novel satellite or sensor systems. It is enough to collect a representative set of input samples to train such unsupervised models, which can speed up the time needed for obtaining automated insight from these missions.

\begin{ack}
This work has been enabled by the Frontier Development Lab Program (FDL). FDL is a collaboration between SETI Institute and Trillium Technologies Inc., in partnership with Department of Energy (DOE), University of Oxford and Planet.
The authors would like to thank all FDL faculty members, Cormac Purcell (University of New South Wales), Ash Hoover and Corbin Kling (Planet), Yarin Gal (University of Oxford) and Chedy Raissi (INRIA) for discussions and comments throughout the development of this work. The authors would also like to thank Carter Katzenberger (Northwest Nazarene University) for his work on the project.

This material is based upon work supported by the Department of Energy [National Nuclear Security Administration] under Award Number DE-AI0000001. Disclaimer: This report was prepared as an account of work sponsored by an agency of the United States Government.  Neither the US Government nor any agency thereof, nor any of their employees makes any warranty, express or implied, or assumes any legal liability or responsibility for the accuracy, completeness or usefulness of any information, apparatus product or process disclosed, or represents that its use would not infringe privately owned rights.  Reference herein to any specific commercial product, process, or service by trade name trademark, manufacturer, or otherwise does not necessarily constitute or imply its endorsement, recommendation, or favoring by the US Government or any agency thereof.  The views and opinions of authors expressed herein do not necessarily state or reflect those of the US Government or any agency thereof.

\end{ack}

\bibliographystyle{unsrtnat}

\bibliography{biblio}

\begin{thebibliography}{16}
\providecommand{\natexlab}[1]{#1}
\providecommand{\url}[1]{\texttt{#1}}
\expandafter\ifx\csname urlstyle\endcsname\relax
  \providecommand{\doi}[1]{doi: #1}\else
  \providecommand{\doi}{doi: \begingroup \urlstyle{rm}\Url}\fi

\bibitem[Bot and Borges(2022)]{bot2022systematic}
Karol Bot and Jos{\'e}~G Borges.
\newblock A systematic review of applications of machine learning techniques
  for wildfire management decision support.
\newblock \emph{Inventions}, 7\penalty0 (1):\penalty0 15, 2022.

\bibitem[Moustakas and Davlias(2021)]{moustakas2021minimal}
Aristides Moustakas and Orestis Davlias.
\newblock Minimal effect of prescribed burning on fire spread rate and
  intensity in savanna ecosystems.
\newblock \emph{Stochastic Environmental Research and Risk Assessment},
  35\penalty0 (4):\penalty0 849--860, 2021.

\bibitem[Khelifi and Mignotte(2020)]{khelifi2020deep}
Lazhar Khelifi and Max Mignotte.
\newblock Deep learning for change detection in remote sensing images:
  Comprehensive review and meta-analysis.
\newblock \emph{Ieee Access}, 8:\penalty0 126385--126400, 2020.

\bibitem[Gallagher et~al.(2021)Gallagher, Maxwell, Guill{\'e}n, Everland,
  Loudermilk, and Skowronski]{gallagher2021estimation}
Michael~R Gallagher, Aaron~E Maxwell, Luis~Andr{\'e}s Guill{\'e}n, Alexis
  Everland, E~Louise Loudermilk, and Nicholas~S Skowronski.
\newblock Estimation of plot-level burn severity using terrestrial laser
  scanning.
\newblock \emph{Remote Sensing}, 13\penalty0 (20):\penalty0 4168, 2021.

\bibitem[Gibson et~al.(2020)Gibson, Danaher, Hehir, and
  Collins]{gibson2020remote}
Rebecca Gibson, Tim Danaher, Warwick Hehir, and Luke Collins.
\newblock A remote sensing approach to mapping fire severity in south-eastern
  australia using sentinel 2 and random forest.
\newblock \emph{Remote Sensing of Environment}, 240:\penalty0 111702, 2020.

\bibitem[Montorio et~al.(2020)Montorio, P{\'e}rez-Cabello, Alves, and
  Garc{\'\i}a-Mart{\'\i}n]{montorio2020unitemporal}
Raquel Montorio, Fernando P{\'e}rez-Cabello, Daniel~Borini Alves, and Alberto
  Garc{\'\i}a-Mart{\'\i}n.
\newblock Unitemporal approach to fire severity mapping using multispectral
  synthetic databases and random forests.
\newblock \emph{Remote Sensing of Environment}, 249:\penalty0 112025, 2020.

\bibitem[Florath and Keller(2022)]{florath2022supervised}
Janine Florath and Sina Keller.
\newblock Supervised machine learning approaches on multispectral remote
  sensing data for a combined detection of fire and burned area.
\newblock \emph{Remote Sensing}, 14\penalty0 (3):\penalty0 657, 2022.

\bibitem[Asokan and Anitha(2019)]{asokan2019change}
Anju Asokan and JJESI Anitha.
\newblock Change detection techniques for remote sensing applications: a
  survey.
\newblock \emph{Earth Science Informatics}, 12\penalty0 (2):\penalty0 143--160,
  2019.

\bibitem[Růžička et~al.(2022)Růžička, Vaughan, De~Martini, Fulton,
  Salvatelli, Bridges, Mateo-Garcia, and Zantedeschi]{ruzicka2021unsupervised}
Vít Růžička, Anna Vaughan, Daniele De~Martini, James Fulton, Valentina
  Salvatelli, Chris Bridges, Gonzalo Mateo-Garcia, and Valentina Zantedeschi.
\newblock Rav{\ae}n: unsupervised change detection of extreme events using ml
  on-board satellites.
\newblock \emph{Scientific reports}, 12\penalty0 (1):\penalty0 1--10, 2022.

\bibitem[Team(2017)]{planet}
Planet Team.
\newblock Planet application program interface: In space for life on earth,
  2017.

\bibitem[Hamilton et~al.(2018)Hamilton, Hamilton, and
  Myers]{hamilton2018evaluation}
Dale Hamilton, Nicholas Hamilton, and Barry Myers.
\newblock Evaluation of image spatial resolution for machine learning mapping
  of wildland fire effects.
\newblock In \emph{Proceedings of SAI Intelligent Systems Conference}, pages
  400--415. Springer, 2018.

\bibitem[Escuin et~al.(2008)Escuin, Navarro, and Fern{\'a}ndez]{escuin2008fire}
S~Escuin, R~Navarro, and P~Fern{\'a}ndez.
\newblock Fire severity assessment by using nbr (normalized burn ratio) and
  ndvi (normalized difference vegetation index) derived from landsat tm/etm
  images.
\newblock \emph{International Journal of Remote Sensing}, 29\penalty0
  (4):\penalty0 1053--1073, 2008.

\bibitem[Wold et~al.(1987)Wold, Esbensen, and Geladi]{wold1987principal}
Svante Wold, Kim Esbensen, and Paul Geladi.
\newblock Principal component analysis.
\newblock \emph{Chemometrics and intelligent laboratory systems}, 2\penalty0
  (1-3):\penalty0 37--52, 1987.

\bibitem[Chen et~al.(2020)Chen, Kornblith, Norouzi, and Hinton]{chen2020simple}
Ting Chen, Simon Kornblith, Mohammad Norouzi, and Geoffrey Hinton.
\newblock A simple framework for contrastive learning of visual
  representations.
\newblock In \emph{International conference on machine learning}, pages
  1597--1607. PMLR, 2020.

\bibitem[He et~al.(2016)He, Zhang, Ren, and Sun]{he2016deep}
Kaiming He, Xiangyu Zhang, Shaoqing Ren, and Jian Sun.
\newblock Deep residual learning for image recognition.
\newblock In \emph{Proceedings of the IEEE conference on computer vision and
  pattern recognition}, pages 770--778, 2016.

\bibitem[Sumbul et~al.(2019)Sumbul, Charfuelan, Demir, and
  Markl]{sumbul2019bigearthnet}
Gencer Sumbul, Marcela Charfuelan, Beg{\"u}m Demir, and Volker Markl.
\newblock Bigearthnet: A large-scale benchmark archive for remote sensing image
  understanding.
\newblock In \emph{IGARSS 2019-2019 IEEE International Geoscience and Remote
  Sensing Symposium}, pages 5901--5904. IEEE, 2019.

\end{thebibliography}

\newpage
\appendix
\section{Appendix}

\subsection{Methodology illustration}

\begin{figure}[h]
    \centering
    \includegraphics[trim=50 30 50 20,clip,width=0.9\textwidth]{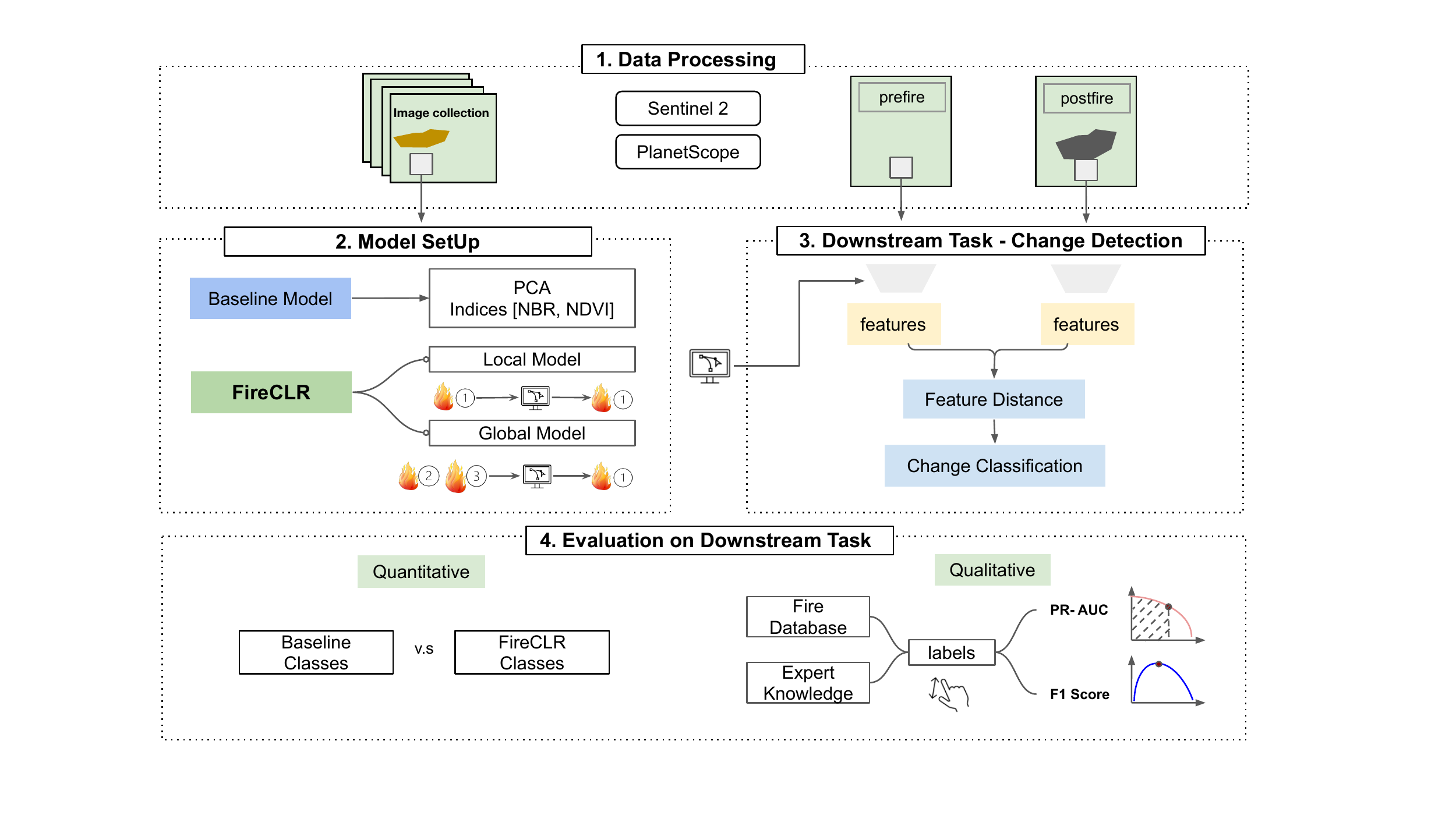}
    \caption{Diagram of Methodology.\label{fig:method}}
\end{figure}

\subsection{Data Acquisition}
\label{sec:data_acq}

PlanetScope imagery from locations mentioned in Table \ref{tab:areas} was obtained in our preliminary study, from which a subset of data was used to train and evaluate models explored in this paper. 
We selected 3 different study areas throughout the Western United States. As long as the imagery covered enough of the study area (>50$\%$) and did not have significant cloud cover, each PlanetScope and Sentinel-2 image was added to the dataset and could be considered either pre-fire, active fire, or post-fire imagery. Data was acquired with these filters across five study areas, as shown in Table \ref{tab:areas}, however, note that only some locations and time frames were used as datasets for the tested methods.

\begin{table}[h]
\caption{\label{tab:areas}Description of Wildfires used in this Research.}

\begin{tabular}{@{}lcccc@{}}
\toprule
\multicolumn{1}{c}{Area of Interest} & Fire Start Date & Containment Date & Time of Interest       & Size (Hectares) \\ \midrule
Mesa Fire                            & 7/26/2018       & 9/25/2018        & 7/1/2018 to 10/1/2018  & 14,050       \\
East Troublesome                     & 10/14/2020      & 11/30/2020       & 10/1/2020 to 12/1/2020 & 78,432      \\
McFarland Fire                       & 7/29/2021       & 9/16/2021        & 7/1/2021 to 10/1/2021  & 49,635      \\ \bottomrule
\end{tabular}
\end{table}
\vspace{-2mm}

The Mesa Fire was located on the Payette National Forest in southern Idaho, the McFarland Fire occurred on the Shasta-Trinity National Forest in northern California and the East Troublesome Fire was located on the Arapaho and Roosevelt National Forests and the Pawnee National Grassland in Colorado.

To account for days where smoke (or other unexpected events) may obstruct the features on the surface of the areas of interest, the time of interest was expanded by two weeks in either direction of the start and end dates. Additionally, the NIR band was not available for every study area, so PlanetScope imagery that included a fourth NIR multispectral band was prioritized over imagery containing only the visual RGB bands.

\subsection{Data preprocessing}
\label{sec:data_prep}

The raw obtained data requires few preprocessing steps before it can be considered as machine learning ready. For PlanetScope data, we first need to mosaic several captures by the satellites into one large stitched image.
Products from Sentinel-2 and PlanetScope data are clipped with a more refined region of interest to correspond to the exact same locations.
All data is first visualised and manually assessed for usage, only scenes without large cloud cover are kept.
Bands in the Sentinel-2 product are all resampled to the same sampling resolution of 10m.
We split the captures from different timeframes and different locations into train and test sets - more about this in the following sections.
Finally, the data is normalised using the statistics from the training set, these parameters are saved and later reused for the rest of the data. We use the global minimum and maximum values to normalise the data.

For selected locations and time frames, namely the \textit{Mesa} fire listed in Table \ref{tab:areas}, we used the fire burn extent as a ground truth label. 
We obtain the label from the National Interagency Fire Center\footnote{\url{https://www.nifc.gov/}}. 
In addition, we also evaluate our model’s performance on detecting the different levels of fire severity, such as its ability to detect the white ash and black ash. We have labeled areas of these two extra classes manually.
We note that these labels were used only for the evaluation of the different model versions and not for training.

\subsection{Evaluation on downstream task}

The goal of the downstream task is to extract meaningful, discriminative features from pre-fire and post-fire images using the model built above. After we have the model trained, we use it to extract the features from pre-fire and post-fire images and calculate the difference between them. The feature distance is calculated both with Euclidean and cosine distance. The K-means clustering is then used to classify the feature distance into multiple classifications such as black ash, white ash and unburned areas.

\subsection{Experimental Setting and Reproducibility}

We used python and the machine learning library pytorch, namely the versions linked to the public repository of the SimCLR codebase\footnote{\url{https://github.com/sthalles/SimCLR}}. As the training environment, we use the virtual machines on the Google Cloud Platform with the \textit{n1-highmem-32} instance equipped with 4 x NVIDIA Tesla K80 GPUs.

To help with the replicability of this study, we are planning to release both the code\footnote{\url{https://github.com/spaceml-org/FireCLR-Wildfires}} describing the used machine learning models, and description of the exact used data. Public part of the data, namely the Sentinel-2 products, will be released with the planned publication of a workshop paper. The data originating from the PlanetScope product cannot be made public, we instead plan to release a list of the used areas and time frames of interest, in order that the data can be identified and obtained by someone with the same access as our team had.

In addition, we are planning to publicly release our models, which were trained in an unsupervised manner on large training datasets from both of the dataset sources. Derived products (such as the trained neural network models) can be made public also for the PlanetScope data. We are planning to release these to further scientific research and we suggest that they could be used as ``out-of-the-box'' pre-trained models similarly as was done for Sentinel-2 data with BigEarthNet \cite{sumbul2019bigearthnet}.


\subsection{Additional downstream experiments}
\label{sec:additional}

\begin{figure}[h]
    \centering
    \includegraphics[width=0.95\textwidth]{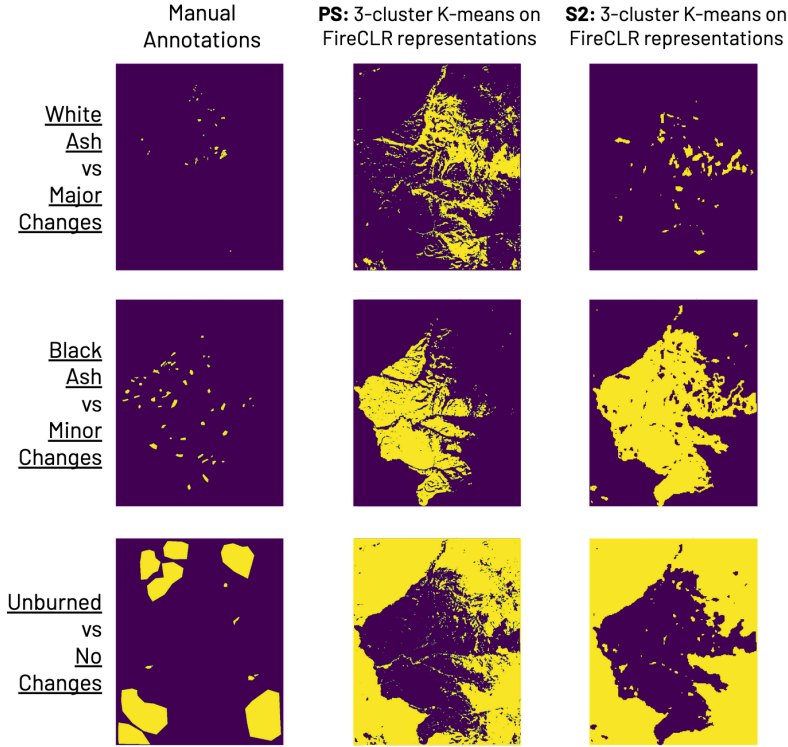}
    \caption{Maps of the downstream experiments. The left column shows the manual annotations by the fire expert, the yellow pixels are the labeled categories. And the second and third columns show the 3-cluster K-means predictions from the latent space representations of the FireCLR models based on the PlanetScope and Sentinel-2 respectively. The yellow pixels are the predicted categories. \label{fig:downstream_maps}}
\end{figure}

The relationships between the magnitudes of the changes based on the PlanetScope and the manual annotations of the burned severity (black and white ashes) were shown in Figure \ref{fig:downstream_maps}. White ash indicates more severely burned areas and corresponds to the major changes. Black ash indicates less severely burned areas and was compared to the minor changes. The major and minor changes were defined based on the K-means clustering of the cosine distance between the pre- and post-fire latent space representations. Based on the qualitative interpretation, the FireCLR model based on the PlanetScope data tends to over predict the white ash and have better performance on predicting the black ash. The FireCLR model based on the Sentinel-2 data tends to under predict the white ash and over predict the black ash. The performance of the two FireCLR models were also quantitatively evaluated using F1-score and compared to the baselines.
We show these results in Table \ref{tab:post_class_sentinel2} and Table \ref{tab:post_class_planetscope}. We note that the proposed FireCLR approach reaches similar performance to the baselines for the \textit{Black ash} class, while it under-performs in the \textit{White ash} class in the Sentinel-2 case. Finally on both datasets, it outperforms the baselines in the case of the background \textit{Unburned} class.

\begin{table}[h]
\caption{\label{tab:post_class_sentinel2}F1-scores of the models trained in local mode (the same location for the training and evaluation set, using different times) with Sentinel-2 data on the downstream task of unsupervised cluster classification.}
\begin{tabular}{@{}llcccc@{}}
\toprule
\multicolumn{1}{c}{Method} & \multicolumn{1}{c}{Bands} & Effective res. & White Ash     & Black Ash     & Unburned      \\ \midrule
PCA + K-means              & S2 RGB+NIR       & 10m                  & 0.59          & \textbf{0.86} & 0.6           \\
dNBR + K-means             & S2 RGB+SWIR      & 10m                  & \textbf{0.93} & 0.78          & 0.76          \\
FireCLR + K-means           & S2 RGB+NIR       & 80m                  & 0.51          & 0.82          & \textbf{0.79} \\ \bottomrule
\end{tabular}
\end{table}

\begin{table}[h]
\caption{\label{tab:post_class_planetscope}F1-scores of the models trained in global mode (different location and timeframes for the training and evaluation sets) with PlanetScope data on the downstream task of unsupervised cluster classification.}
\begin{tabular}{@{}llcccc@{}}
\toprule
\multicolumn{1}{c}{Method} & \multicolumn{1}{c}{Bands} & Effective res. & White Ash    & Black Ash     & Unburned      \\ \midrule
PCA + K-means              & PS RGB+NIR                & 3m             & \textbf{0.9} & \textbf{0.86} & 0.76          \\
FireCLR + K-means           & PS RGB+NIR                & 24m            & \textbf{0.9} & \textbf{0.86} & \textbf{0.78} \\ \bottomrule
\end{tabular}
\end{table}
\end{document}